\pdfoutput=1

\documentclass[11pt]{article}

\usepackage[final]{acl}

\usepackage{times}
\usepackage{latexsym}

\usepackage[T1]{fontenc}

\usepackage[utf8]{inputenc}

\usepackage{microtype}

\usepackage{inconsolata}

\usepackage{graphicx}

\usepackage{enumitem}
\setlist{noitemsep}
\usepackage{mdframed}
\usepackage{fancyvrb}
\usepackage{tcolorbox} 
\usepackage{listings}
\usepackage{multirow}
\DeclareUnicodeCharacter{2060}{}
\usepackage{booktabs}

\usepackage{longtable}

\newmdenv[
  linecolor=blue,
  linewidth=2pt,
  backgroundcolor=blue!10,
  roundcorner=5pt,
  skipabove=10pt,
  skipbelow=10pt
]{examplebox}

\title{HalluDetect: Detecting, Mitigating, and Benchmarking Hallucinations in Conversational Systems in the Legal Domain}

\author{Spandan Anaokar\textsuperscript{\rm 1}, \ Shrey Ganatra\textsuperscript{\rm 1}, \ Harshvivek Kashid\textsuperscript{\rm 1}, \ Swapnil Bhattacharyya\textsuperscript{\rm 1}, \\
\textbf{Shruti Nair}\textsuperscript{\rm 2}, \ \textbf{Reshma Sekhar}\textsuperscript{\rm 2}, \ 
\textbf{Siddharth Manohar}\textsuperscript{\rm 2}, \ 
\textbf{⁠Rahul Hemrajani}\textsuperscript{\rm 2}, \\ 
\textbf{Pushpak Bhattacharyya}\textsuperscript{\rm 1} \\
\textsuperscript{\rm 1}Indian Institute of Technology Bombay\\
\textsuperscript{\rm 2}National Law School of India University, Bangalore\\
\texttt{\{spandananao, ganatrashrey2002, harshvivek14, pushpakbh\}@gmail.com}\\
\texttt{swapnilbhyya@cse.iitb.ac.in}
}

\begin{document}
\maketitle
\begin{abstract}
Large Language Models (LLMs) are widely used in industry but remain prone to hallucinations, limiting their reliability in critical applications. This work addresses hallucination reduction in consumer grievance chatbots built using LLaMA 3.1 8B Instruct, a compact model frequently used in industry. We develop \textbf{HalluDetect}, an LLM-based hallucination detection system that achieves an F1 score of \textbf{68.92\%} outperforming baseline detectors by \textbf{22.47\%}. Benchmarking five hallucination mitigation architectures, we find that out of them, AgentBot minimizes hallucinations to \textbf{0.4159} per turn while maintaining the highest token accuracy (\textbf{96.13\%}), making it the most effective mitigation strategy. Our findings provide a scalable framework for hallucination mitigation, demonstrating that optimized inference strategies can significantly improve factual accuracy.\footnote{The code and dataset are publicly available \url{https://github.com/Spandan2003/halludetect}.}
\end{abstract}

\section{Introduction}
Large Language Models (LLMs) like GPT-4 \cite{achiam2023gpt} and Llama-3 \cite{dubey2024llama} have found widespread use in various domains, ranging from customer support to health care and education. However, despite their advantages, LLMs often produce hallucinations. This phenomenon poses a serious challenge in real-world applications.

Hallucination in natural language generation (NLG) refers to the generation of content that is not supported by the input data or context, often resulting in outputs that are factually incorrect, irrelevant, or inconsistent \cite{Huang_Survey}. Unlike traditional errors that may be easy to spot, hallucinations are often subtle and difficult to detect, requiring more than just surface-level checks.

In critical domains such as legal, medical, and financial services, the accuracy of LLM-generated content is essential. A chatbot providing incorrect legal guidance may mislead consumers into taking actions that could have legal or financial consequences \cite{legal_financial_risks}. Similarly, errors in medical chatbots, such as incorrect medication dosages, can lead to severe health risks \cite{healthcare_risks}. In customer support, hallucinated responses can erode trust and damage a company’s reputation \cite{trust_damage}. The risks associated with hallucinations highlight the urgent need for effective detection and mitigation strategies to ensure reliability and compliance with ethical and regulatory standards, such as the EU AI Act \cite{eu_ai_act}.

In this work, we focus on hallucination in the context of consumer law chatbots \cite{grahaknyay} in India. Legal information is often complex, and consumers rely on automated systems to guide them in resolving disputes, filing complaints, and understanding their rights. Hallucinations in this domain can misinform users, leading to delays or incorrect legal actions. To address this, we explore various hallucination mitigation techniques tailored to consumer law, develop a hallucination detection system, and benchmark the effectiveness of different mitigation methods. Additionally, we carry out a human evaluation to assess the detector’s reliability, complemented by a qualitative analysis of its failure instances.

Our contributions are: 
\begin{enumerate}
    \item \textbf{RAG-based Chatbots}, a collection of consumer grievance chatbots incorporating multiple hallucination mitigation strategies named Vanilla, Prompt-engineered, FactChecker, EditorBot, and AgentBot (Section \ref{sec:ConsumerGrievanceChatbots}).  
    \item \textbf{HalluDetect}, an LLM-driven multi-turn hallucination detection pipeline for scalable evaluation of chatbot conversations, achieving F1 of \textbf{68.92\%}. It identifies specific hallucinated text with justifications and outperforms baseline detectors by \textbf{22.47\%} (Section \ref{sec:HalluDetect}).  
    \item  \textbf{DetectorEval Dataset}, a benchmarking dataset consisting of 115 chats of 7 turns and 1282 tokens on average between the Vanilla chatbot and legal experts, annotated for hallucination detection by human annotators.  (Section \ref{sec:DetectorEval Dataset}).
    \item \textbf{Architectural Benchmarking}, a comparative hallucination analysis across Consumer Grievance Chatbots using HalluDetect, quantifying the effectiveness of different hallucination mitigation strategies. Results show that AgentBot performs best with \textbf{0.4159} hallucinations per conversation turn and the highest Token Accuracy (TokAcc-1: 96.13\%, TokAcc-2: 96.38\%) (Section \ref{sec:Human Evaluation}).
\end{enumerate}
By addressing hallucination detection, mitigation, and benchmarking in consumer law chatbots, our work aims to enhance the safety, trustworthiness, and effectiveness of AI-driven legal assistance tools.

\section{Related Work}

\begin{figure}
\centering
    \tcbset{colframe=blue,colback=blue!10, boxrule=0.5mm, arc=0mm}
    \begin{tcolorbox}
    \small
Covering medical expenses is a reasonable expectation from a service provider like a tattoo parlour.

As a first step, I would recommend drafting a formal complaint letter to the tattoo parlour, stating the details of the incident, the harm caused, and the relief you're seeking. \textcolor{red}{\textbf{I can provide you with a sample draft if you'd like}.} 

Additionally, you can also file a complaint with the National Consumer Helpline (1800-11-4000) for immediate assistance.

\textcolor{red}{\textbf{Before we proceed, I have one more question: Do you have any evidence or documentation to support your claim?}}
    \end{tcolorbox}
    \caption{Example of erroneous hallucination detection by LettuceDetect. Red-highlighted polite language and clarifying questions are incorrectly flagged as hallucinations, revealing the model’s difficulty in distinguishing conversational tone from factual inaccuracy.}
    \label{fig:lettuce-detect}
\end{figure}

Hallucination detection in LLMs has been approached through external retrieval~\cite{chen2024complex}, self-correction~\cite{gou2024critic}, and fine-grained metrics like FactScore~\cite{min2023factscore}. Retrieval-augmented generation (RAG) methods~\cite{huo2023retrieving} and tool-based frameworks like FacTool~\cite{chern2023factool} improve factuality in tasks such as summarization and QA.

However, most approaches focus on single-turn, static contexts and underperform in multi-turn dialogue, where hallucinations arise from long context windows, implicit assumptions, or shifting user intent~\cite{park2025mitigating}. Even RAG systems can produce hallucinations despite correct retrieval~\cite{sun2025redeepdetectinghallucinationretrievalaugmented}, and partial hallucinations remain hard to detect~\cite{budzianowski2018multiwoz}.

Recent detectors—LettuceDetect~\cite{kovács2025lettucedetecthallucinationdetectionframework}, HHEM v2~\cite{hhem-2.1-open}, SelfCheckGPT~\cite{manakul2023selfcheckgptzeroresourceblackboxhallucination}, and RefChecker~\cite{hu2024refcheckerreferencebasedfinegrainedhallucination}—offer scalable, interpretable solutions but are limited to single-turn or document-level contexts and often mislabel benign or inferential responses as hallucinations. An example of LettuceDetect failing is provided in Figure \ref{fig:lettuce-detect}.

To overcome these gaps, we present \textbf{HalluDetect}, a multi-turn, task-oriented detection framework for RAG-based conversations. It handles context-aware inconsistencies, assigns reason codes, and filters pragmatically valid content, making it well-suited for sensitive domains like legal and consumer grievance assistance.

\section{Methodology}

\label{sec:Methodology}
\begin{figure*}[t]
    \centering
    \includegraphics[width=0.95\linewidth]{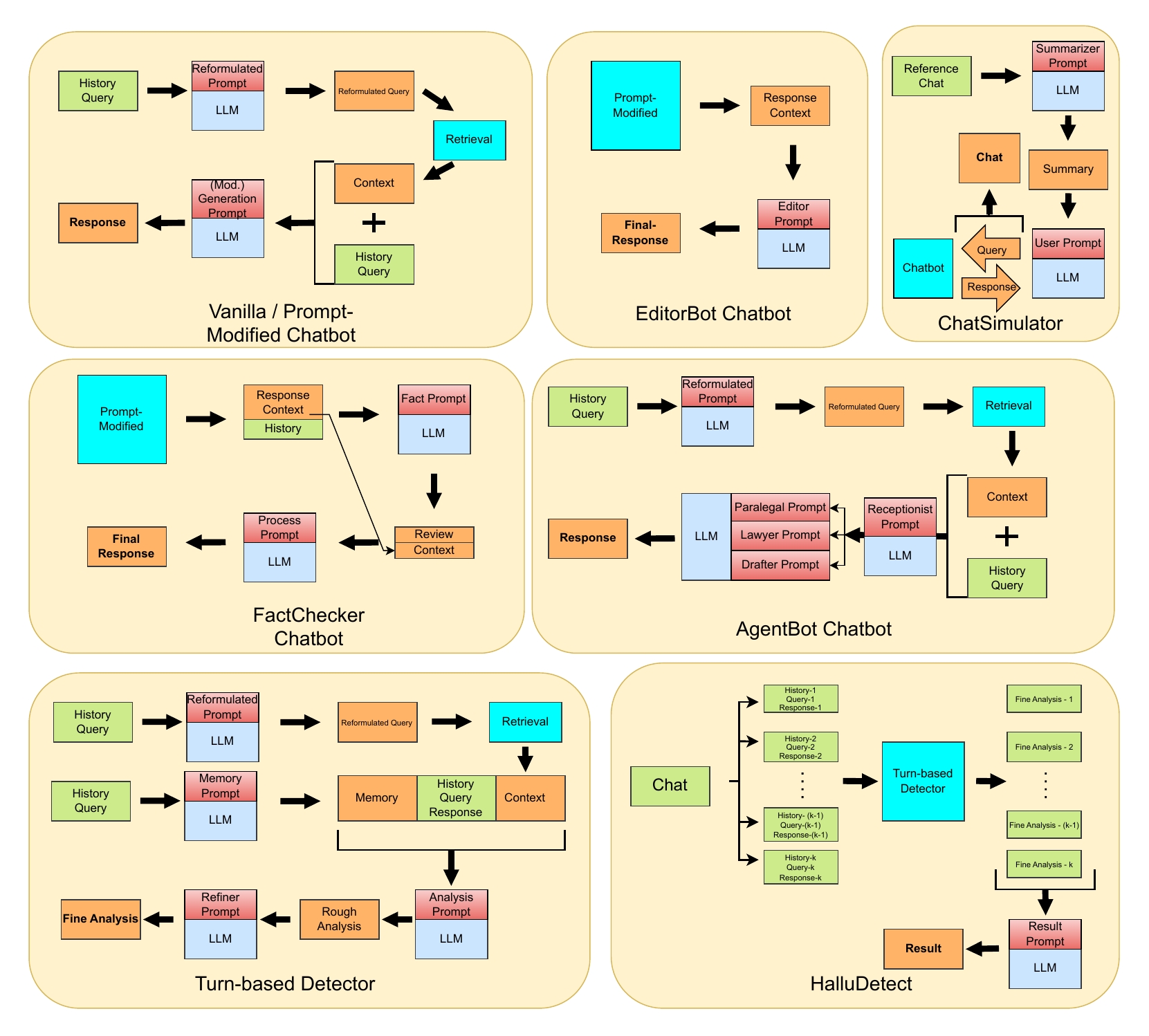}
    \caption{Overview of the architectures for Chatbots, ChatSimulator, and HalluDetect. Each system integrates LLM modules that process specific inputs (green boxes) through prompts (red boxes) to generate outputs (orange boxes)}
    \label{fig:architecture}
\end{figure*}

\subsection{RAG based Chatbots}
\label{sec:ConsumerGrievanceChatbots}
We design a suite of chatbot architectures based on Retrieval-Augmented Generation (RAG) to reduce hallucinations in the legal domain, leveraging RAG’s strength in minimizing factual errors in knowledge-intensive tasks. The chatbots integrate with a curated legal knowledge base covering over 20 consumer sectors in India, such as banking, telecom, and healthcare. Relevant legal documents are retrieved to inform responses. Before detailing each variant, we define key operational components:

\begin{itemize}

    \item \textbf{History:} The record of prior interactions between the user and the chatbot.
    \item \textbf{Query:} The user's latest input, constituting the question or statement requiring a response.
    \item \textbf{Reformulated} \textbf{Query:} A context-independent version of the query generated by integrating the query with its history. \cite{shuster2021retrievalaugmentation}
    \item \textbf{Context:} The set of relevant legal documents or textual knowledge retrieved based on the reformulated query. Also assumed to be the ground truth
    \item \textbf{Response:} The output generated by the chatbot synthesises information from the query, history, and retrieved context.
\end{itemize}

We evaluate five chatbot architectures designed to reduce hallucinations by integrating retrieval and verification mechanisms:  
\begin{itemize}  
    \item \textbf{Vanilla (based on \cite{grahaknyay}):} A baseline RAG pipeline using two LLM calls—one for query reformulation and one for response generation. It does not include explicit hallucination control mechanisms.
    
    \item \textbf{Prompt-engineered:} Builds on Vanilla by refining the response-generation prompt to explicitly instruct the model to stay within the retrieved context and avoid speculation. This improves factual alignment via prompt optimization.
    
    \item \textbf{EditorBot (inspired by \cite{gou2024critic}):} Adds a third LLM call to post-process and revise the generated response. This step aims to identify and remove hallucinations using an \textit{editor prompt} that critiques the initial output.
    
    \item \textbf{FactChecker (inspired by \cite{min2023factscore}):} Decomposes the response into discrete factual claims, verifies each against the retrieved context using a second LLM, and synthesizes a revised, verified response. This adds structure to fact-checking and involves three LLM calls in total.
    
    \item \textbf{AgentBot (inspired by \cite{kwartler2024goodparentingneed}):} Uses a multi-agent workflow mimicking real-world legal processes. Distinct roles—Receptionist, Paralegal, Lawyer, Drafter—each handle a specific aspect of the task. This structured decomposition encourages specialization and improves factual consistency.
 
\end{itemize}  
The overall architecture is given in the form of an illustration in Figure~\ref{fig:architecture}, with complete prompt details provided in Appendix~\ref{sec:architecture_details} for reference.

\subsection{HalluDetect: LLM-Based Evaluation}
\label{sec:HalluDetect}

\textbf{Overview.}
HalluDetect is a multi-stage LLM pipeline for detecting hallucinations in multi-turn chatbot conversations. It is designed for RAG-based systems where factual grounding may come from both retrieved documents and the LLM's pretraining. This balance is critical in domains like law, where not all information is explicitly retrievable.

\textbf{Domain Reliance Paradox.}
In RAG setups, verifying responses against retrieved documents alone can misclassify legitimate general knowledge as hallucinated (Fig. \ref{fig:lettuce-detect}), while under-relying may allow real hallucinations to pass. HalluDetect addresses this trade-off by combining expanded context, memory summarization, and a severity-based filtering mechanism.

\textbf{Pipeline Structure.}
The detector operates on (history, query, response) triples and proceeds in three stages:
\begin{enumerate}
    \item \textbf{Analysis:} Retrieves twice the original documents and uses a \textit{memory\_prompt} to summarize the dialogue history. An LLM identifies potential hallucinations based on this context.
    \item \textbf{Filtering:} A \textit{refiner\_prompt} assigns severity scores (1–5) and filters out low-impact hallucinations (score < 4).
    \item \textbf{Aggregation:} High-severity hallucinations are compiled as final output, each with justification and location in the dialogue.
\end{enumerate}

\textbf{Design Features.}
Key features include:
(i) doubling the retrieved document number to reduce false positives, 
(ii) memory created by summarizing long contexts, 
(iii) severity scoring to filter trivial or inaccurate content, and 
(iv) justification-based self-consistency for robust detection~\cite{wang2023selfconsistencyimproveschainthought}.

\textbf{Output Format.}
Each hallucination in the final output includes: (i) the exact sentence suspected to be incorrect, (ii) a justification referencing the retrieved or missing evidence, and (iii) dependent on its place in the multi-turn conversation, ensuring appropriate context is considered.

\textbf{Scalability and Application.}
HalluDetect is domain-agnostic and can audit hallucinations across different chatbot architectures. Its structured format can ensure consistent evaluation in other sensitive domains, such as legal or medical assistance. Implementation details are in Appendix~\ref{sec:hallu_detect_details}.

\subsection{DetectorEval Dataset}
\label{sec:DetectorEval Dataset}
To evaluate the effectiveness of hallucination detection models, we introduce the \textbf{DetectorEval} dataset. It contains 115 dialogue instances between the Vanilla chatbot and legal experts, each crafted to target specific, nuanced issues within the consumer grievance domain. These conversations are annotated by experienced annotators (See Appendix ~\ref{sec:detectoreval-annotation}) using HalluDetect to flag hallucinations. On average, each chat consists of 7.39 turns and 4.04 detected hallucinations. Each turn contains an average of 223.56 tokens, with an average of 1282.6 tokens per conversation. 

\section{Experiments}

\subsection{Structure of Chatbot}
To enable retrieval-augmented generation (RAG), the legal corpus is segmented into coherent text \textit{chunks} (e.g., paragraphs), which are embedded using the \texttt{mixedbread-ai/mxbai-embed-large-v1}\footnote{\url{https://huggingface.co/mixedbread-ai/mxbai-embed-large-v1}} model and stored in a vector store. At inference, the retriever computes cosine similarity between the query and all stored chunks, selecting the top four as supporting context. These chunks guide grounded response generation. We use Llama-3.1-8B-Instruct\footnote{\url{https://huggingface.co/meta-llama/Llama-3.1-8B-Instruct}} and GPT-4o mini\footnote{\url{https://platform.openai.com/docs/models/gpt-4o-mini}} as the response generators.

\subsection{Human Evaluation of HalluDetect}
\label{sec:Human Evaluation}
We evaluate HalluDetect on the DetectorEval dataset, where it extracts hallucinated content from multi-turn chats. To ensure completeness, language experts manually review its outputs, identifying both correctly detected and missed hallucinations.

\textbf{Metrics:} Annotators label detected hallucinations as Correct or Wrong and report any missed instances. Precision is the fraction of correct detections among all flagged cases; recall is the fraction of correct detections among all actual hallucinations. We report average precision and recall across conversations. Detailed instructions are in Appendix  (\ref{lst:instruction})

\textbf{Baselines:} HalluDetect is compared with two SOTA RAG-focused detectors:
\begin{enumerate}
    \item \textbf{HHEM 2.1}~\cite{hhem-2.1-open}: A single-turn detector for RAG combining heuristic and NLI-based scoring (Flan-T5).
    \item \textbf{LettuceDetect}~\cite{kovács2025lettucedetecthallucinationdetectionframework}: A single-turn detector using token-level entropy and fact verification via \texttt{base-modernbert-en-v1}.
\end{enumerate}
Each baseline produces turn-wise hallucination labels for comparison.

    

\subsection{Comparative Analysis of Chatbot Architectures}
To evaluate hallucination mitigation strategies, we use the \textbf{ChatSimulator} (Figure~\ref{fig:architecture}), which interfaces with different Consumer Grievance Chatbot variants. Instead of a human, it uses a separate LLM agent with guided instructions to simulate realistic conversations. Summaries from 30 DetectorEval chats provide contextual grounding for consistent simulation. Each generated dialogue is evaluated by HalluDetect to detect and rate hallucinations, enabling fair comparison across architectures.

\textbf{Metrics:} We report Hallucinations per Turn (HPT), the average number of hallucinations generated per chatbot turn, and Token Accuracy (TokAcc), the percentage of non-hallucinatory tokens in responses. HPT-1 and TokAcc-1 are computed over the entire dataset, while HPT-2 and TokAcc-2 are averaged per conversation. These metrics enable standardized comparison of hallucination frequency and response reliability.

\section{Results and Analysis}

\subsection{Quantitative Performance Analysis}
We first evaluate HalluDetect on the Vanilla chatbot outputs from the DetectEval dataset, then apply it on ChatSimulator generated dialogues to assess hallucination mitigation strategies across chatbot variants.

\subsubsection{Detector Evaluation}
HalluDetect achieves an average precision of 54.56\% and recall of 80.63\% (Table~\ref{tab:human_eval}), indicating high coverage of hallucinations with reasonable accuracy. Its strong recall ensures most hallucinations are detected, while precision reflects the detector’s ability to avoid false positives, making it reliable for downstream analysis.


\begin{table}[ht]
    \centering
    \begin{tabular}{|p{2.2cm}|c|c|c|}
        \hline
        \textbf{Model} & \textbf{Prec.} & \textbf{Recall} & \textbf{F1}  \\ 
        \hline
        LettuceDetect (Baseline) & 0.3055 & \underline{0.9689} & 0.4645\\ 
        \hline
        HHEM2.1 (Baseline) & 0.2791 &  \textbf{0.9835} & 0.4348\\ 
        \hline
        HalluDetect (Llama 3.1) &  \textbf{0.5456} & 0.8063 & \underline{0.6508}\\ 
        \hline
        HalluDetect (gpt-4o-mini) & \underline{0.5430} & 0.9429 & \textbf{0.6892}\\ 
        \hline
    \end{tabular}
    \caption{Human Evaluation of HalluDetect and Baseline Models.}
    \label{tab:human_eval}
\end{table}

We compare HalluDetect with baseline models, LettuceDetect and HHEM2.1. While all models show high recall, the baselines suffer from significantly lower precision. All are evaluated on the full DetectorEval set (Section~\ref{sec:DetectorEval Dataset}). The precision drop stems from chatbot responses often including suggestions, questions, or common-sense reasoning (e.g., “the doctor should have given the correct medicines”), which baselines—designed for single-turn QA—misclassify as hallucinations. HalluDetect addresses this by filtering such benign or inferred content via its multistage pipeline, achieving substantially higher precision with only a minor recall tradeoff.

HalluDetect also adds only a modest computational overhead-about 2.52× the base chatbot’s token usage-since runtime scales linearly with tokens rather than model complexity. A detailed efficiency breakdown is provided in Appendix~\ref{sec:efficiency_appendix}.

\subsubsection{Statistical Significance Testing}

To confirm that HalluDetect’s improvements were not due to random variation, we conducted a Friedman test across detectors followed by pairwise Wilcoxon signed-rank tests between HalluDetect and the other baselines. As shown in Table~\ref{tab:statsig}, the Friedman test revealed statistically significant differences in both precision and recall ($p < 0.01$). Pairwise comparisons further confirm that HalluDetect achieves significantly higher precision than HHEM and LettuceDetect ($p < 0.001$), while maintaining a statistically comparable but slightly lower recall. These results validate that HalluDetect’s performance gains are consistent across samples.

\begin{table}[h]
\centering
\small
\begin{tabular}{|p{1.4in}|p{0.5in}|p{0.6in}|}
\hline
\textbf{Metric} & \textbf{Test / Comparison} & \textbf{$p$-value} \\
\hline
\textit{\textbf{Friedman Test} (Overall)} &  $\mathbf{\chi^2}$ &\\
Precision & 26.77 & $2\times10^{-6}$\\
Recall    &  9.52  & $8.6\times10^{-3}$ \\
\hline
\textit{\textbf{Wilcoxon Tests} (HalluDetect)}& $\mathbf{W}$ &\\
Precision (vs HHEM) & $445.0$ & $1\times10^{-6}$\\
Precision (vs LettuceDetect) & $494.0$ & $2\times10^{-5}$\\
Recall (vs HHEM) & $51.0$ & $1.6\times10^{-3}$\\
Recall (vs LettuceDetect) & $70.0$ & $1.3\times10^{-2}$\\
\hline
\end{tabular}
\caption{Statistical significance tests for detector performance. All $p$-values below 0.05 indicate significant differences.}
\label{tab:statsig}
\end{table}

\begin{table*}[t]
    \centering
    \begin{tabular}{|p{0.7in}|c|p{0.7in}|c|c|c|c|p{0.7in}|}
        \hline
        {\centering\textbf{LLM}} & \textbf{Variant} & \textbf{Inferences/ Response} & \textbf{HPT1} & \textbf{HPT2} & \textbf{TokAcc-1} & \textbf{TokAcc-2} & \textbf{\% of Hallucinated Chats}\\
        \hline
        \multirow{5}{*}{\shortstack{LLaMA-3.1\\-8B-Instruct}} & Vanilla & 2 & 0.8165 & 0.7901 & 0.9383 & 0.9436 & 60.00\%\\
        \cline{2-8}
        & Prompt-Eng. & 2 & 0.7692 & 0.8944 & 0.9400 & 0.9343 & 63.33\%\\
        \cline{2-8}
        & EditorBot & 3 & \underline{0.4476} & \underline{0.4744} & 0.9564 & 0.9547 & 56.67\%\\
        \cline{2-8}
        & FactChecker & 4 & 0.5872 & 0.5861 & \underline{0.9589} & \underline{0.9597} & \underline{50.00\%}\\
        \cline{2-8}
        & AgentBot & 3 & \textbf{0.4159} & \textbf{0.4279} & \textbf{0.9613} & \textbf{0.9638} & \textbf{46.67}\%\\
        \hline
    \end{tabular}
    \caption{Comparison of different Chatbot Architectures for Hallucination Mitigation. We can see that AgentBot gives the least hallucination due to its low HPT and high TokAcc.}
    \label{tab:combined_llm_comparison}
\end{table*}

\subsubsection{Evaluation of Chatbots using HalluDetect}

AgentBot achieves the best performance, with the lowest Hallucinations per Turn (HPT1: 0.42, HPT2: 0.43) and highest Token Accuracy (TokAcc-1: 96.13\%, TokAcc-2: 96.38\%), and only 46.67\% of chats containing hallucinations. EditorBot follows with slightly lower accuracy (95.64\%, 95.47\%) and marginally higher HPT (0.45, 0.47), while FactChecker performs comparably. In contrast, Vanilla and Prompt-engineered bots show significantly higher hallucination rates. Thus, \textbf{AgentBot} strikes the best balance between accuracy and hallucination control.

\subsubsection{ChatSimulator}
This section compares chatbot architectures based on turns and tokens generated in the ChatSimulator, offering insights into verbosity and consistency. AgentBot generates the longest conversations (999 tokens on average), with high variability, suggesting verbose responses but low consistency. In contrast, Prompt-Engineered and EditorBot show more consistent, concise patterns with lower variability. These differences highlight the impact of architectural choices on verbosity and consistency, crucial for applications requiring structured dialogue, like customer service or personal assistants.

\begin{figure}[h] 
    \centering
    \includegraphics[width=0.45\textwidth]{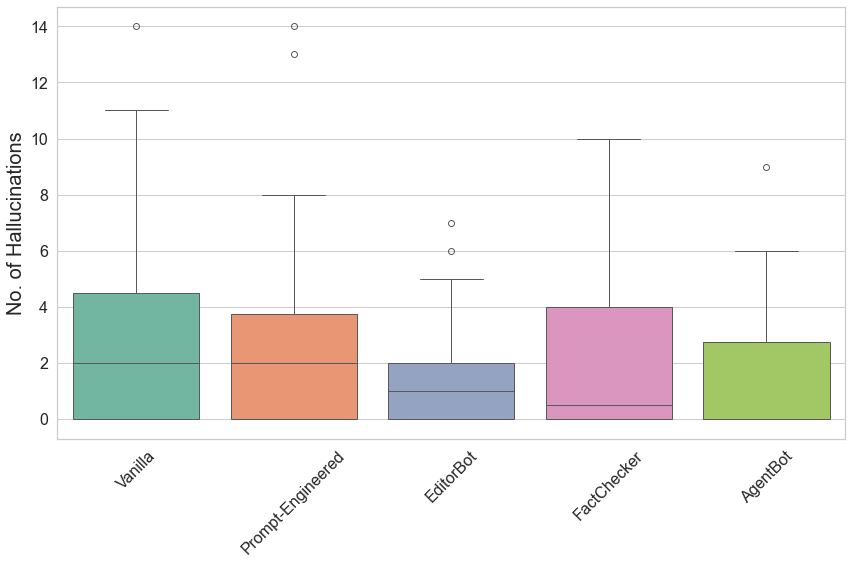}  
    \caption{Box Plot of Count of Hallucinations detected by HalluDetect per Chat for each chatbot variant.} 
    \label{fig:box_plot} 
\end{figure}

\subsection{Qualitative Analysis}

To complement quantitative evaluation, we present a qualitative analysis to illustrate HalluDetect’s effectiveness in identifying and filtering hallucinations in multi-turn conversations. This analysis highlights its strengths in recognizing substantive factual inconsistencies while avoiding penalization of benign or inferential statements.

\paragraph{Example:}
In a conversation about consumer courts in Tripura, the chatbot provides both factual and conversational content, including details about a state commission, contact numbers, and a closing offer to help.

HalluDetect identifies three hallucinated spans in the initial analysis phase:
\begin{enumerate}
\item \textbf{“State Consumer Dispute Redressal Commission, Tripura, Agartala.”} — Not present in retrieved context. \textit{(Severity: 4)}
\item \textbf{“...0381-232-0325 or 0381-232-0326.”} — Unsupported phone numbers. \textit{(Severity: 5)}
\item \textbf{“Feel free to ask...”} — Polite offer not grounded in retrieval. \textit{(Severity: 1)}
\end{enumerate}

HalluDetect processes this conversational turn using the Analysis Prompt, which generates a Rough Analysis—a list of extracted hallucinated spans along with explanations for why they are considered hallucinations. This output is then passed through the Refiner Prompt, producing the Fine Analysis, where each hallucination is assigned a severity score and justification. Low-severity items, such as point 3 (a polite, general offer by the LLM), are filtered out at this stage. This process is repeated for each turn, and the high-severity results are aggregated to yield the final hallucination detections for the complete dialogue. Complete conversational turn details are provided in Appendix (Section \ref{sec:appendix_example})

\subsubsection{Failure Case Analysis}
To better understand the limitations of HalluDetect, we conduct a qualitative analysis of representative failure cases. These include both \emph{false positives}, where the detector incorrectly flags grounded information as hallucinated, and \emph{false negatives}, where genuine hallucinations escape detection. This analysis helps clarify the current boundaries of our approach and guides potential directions for refinement.

\paragraph{False Positive.}
\begin{quote}
\small
\textbf{AI:} Here's a sample notice template you can use as a starting point.\\
\textbf{[Notice to Sharan Hegde]}\\
\textbf{Email:} sharan.hegde@gmail.com
\end{quote}

\textbf{Description:} The detector flagged the email address as hallucinated since it was not found in the RAG corpus.  
\textbf{Analysis:} In this case, the email was part of the chat history provided by the user, but not explicitly retrieved from the external knowledge base. This indicates that the model occasionally fails to integrate conversational context into its grounding check, leading to context-based false positives.

\paragraph{False Negative.}
\begin{quote}
\small
\textbf{AI:} You can file a complaint at the Air India office in Bangalore located at \textit{No. 32, 4th Floor, Brigade Gateway, 26/27, Off MG Road, Bangalore 560025.}
\end{quote}

\textbf{Description:} The detector did not flag this response, though the address is factually incorrect and absent from the RAG corpus.  
\textbf{Analysis:} The hallucination went undetected because the response was fluent and appeared semantically coherent. HalluDetect's current mechanism relies primarily on factual overlap and entity-level consistency; however, sometimes such cases pass undetected, especially if it relies on its pretrained knowledge rather than the actual RAG context.

\paragraph{Discussion.}
These examples highlight that most detection failures fall into two categories:
(i) contextual confusion from prior chat history (false positives), and  
(ii) subtle factual hallucinations phrased naturally (false negatives).  
Despite these limitations, HalluDetect remains robust across diverse dialogue settings and provides interpretable, consistent signals of hallucination risk.

\section{Conclusion and Future Work}
This paper evaluates hallucination detection in chatbots using the HalluDetect framework, applied to multiple chatbot architectures tested through a ChatSimulator. Results show that all hallucination mitigation strategies outperformed the Vanilla model in reducing hallucinations. HalluDetect provides much better precision and slightly lower recall, compared to other SOTA baselines, proving its effectiveness for multi-turn chatbots.

HalluDetect provides an automated, real-time solution for detecting hallucinations in chatbots, which can be used for ongoing evaluation and improvement. This method is scalable and can be integrated into real-time systems to monitor chatbot performance and flag hallucinations as they occur, offering significant benefits for industries such as customer service, healthcare, and legal assistance where factual accuracy is critical.


\section*{Limitations}
Our proposed framework inherits certain limitations common to LLM-based systems. As our detector itself is built on a language model, it may occasionally hallucinate or misclassify outputs, particularly in ambiguous or borderline cases. Furthermore, although the framework is designed to be domain-agnostic, our evaluations are currently limited to the legal consumer grievance domain. Its effectiveness in other domains, while theoretically feasible, has not yet been empirically validated.

\section*{Ethics Statement}

This work adheres to the ACL Code of Ethics. The HalluDetect framework is developed to enhance the factual reliability of large language models (LLMs) in sensitive domains such as legal assistance and consumer grievance redressal. All datasets used—including DetectorEval—contain synthetic or anonymized dialogues, ensuring no personally identifiable information is present. Human evaluation was conducted by domain experts with appropriate annotation guidelines to ensure fair and accurate assessments. Our system is designed for audit and analysis purposes, not for direct legal advice or decision-making. We acknowledge that even LLM-based detectors can introduce errors or biases, and we advocate responsible deployment with human oversight in real-world applications.

\section*{Acknowledgements}

We dedicate this work to the memory of Prof. Pushpak Bhattacharyya, our guide, whose guidance and encouragement were integral to the success of this project. We sincerely thank the legal experts at the National Law School of India University, Bangalore, for their invaluable assistance in curating and annotating the dataset. We also acknowledge the contributions of the Computation for Indian Language Technology Lab (CFILT), IIT Bombay, for their significant support in evaluating the model. Finally, we are grateful to META for their generous funding, which made this project possible.

\bibliography{custom}

\appendix
\newpage

\section{Detailed Model Architectures}

\label{sec:architecture_details}  
\subsection{Consumer Grievance Chatbots}
For a high-level overview, please refer back to Section~\hyperref[sec:ConsumerGrievanceChatbots]{\ref*{sec:ConsumerGrievanceChatbots}}.
\subsubsection{Vanilla Model}  
The baseline model follows a two-step RAG pipeline:  
1. The user query, concatenated with conversation history, is processed by the LLM via the \textit{reformulation\_prompt} to generate a refined query.  
2. This reformulated query retrieves the most relevant legal documents from the knowledge base.  
3. The retrieved documents, along with the original query and history, are passed into the LLM with the \textit{generation\_prompt}, producing the final response.  

\subsubsection{Prompt-engineered Model}  
This variant builds upon the Vanilla Model by optimizing the \textit{generation\_prompt} to explicitly enforce factual correctness. The refinements are based on analyzing common hallucination patterns in the Vanilla Model’s outputs. Additional factual grounding cues, such as explicit retrieval references and conditional response structures, are incorporated to improve accuracy.  

\subsubsection{EditorBot}  
Inspired by \cite{gou2024critic}, EditorBot introduces a post-generation verification step:  
1. The response is first generated using the Prompt-engineered Model.  
2. The response is then evaluated by an LLM using the \textit{editor\_prompt}, which is explicitly designed to detect and correct factual inconsistencies.  
3. The final response, verified for alignment with retrieved legal documents, is then presented to the user.  
This additional inference step enhances factual correctness while preserving the original response’s structure.  

\subsubsection{FactChecker Model}  
Based on \cite{min2023factscore}, this model follows a structured fact verification process:  
1. The initial response is generated using the Prompt-engineered Model.  
2. The response is then segmented into factual claims and miscellaneous statements using an LLM and the \textit{fact\_prompt}.  
3. Each extracted factual claim is independently verified using retrieval-based evidence. If a claim is unsupported, it is flagged as a hallucination.  
4. A final LLM pass, using the \textit{process\_prompt}, refines the response by modifying or removing unsupported claims before outputting the final version.  

\subsubsection{AgentBot}  
Inspired by \cite{kwartler2024goodparentingneed}, AgentBot modularizes the chatbot’s response pipeline into specialized sub-agents:  
\begin{itemize}  
    \item \textbf{Receptionist:} Determines the type of user query and routes it to the appropriate sub-agent.  
    \item \textbf{Paralegal:} Retrieves relevant legal precedents and documents.  
    \item \textbf{Lawyer:} Analyzes legal context and formulates an accurate response.  
    \item \textbf{Drafter:} If required, generates formal legal documents, ensuring compliance with domain-specific guidelines.  
\end{itemize}  
By breaking down response generation into task-specific inferences, AgentBot enhances accuracy while maintaining domain-specific consistency.  

\section{HalluDetect: Detailed Evaluation Workflow}
\label{sec:hallu_detect_details}

HalluDetect is a structured multi-stage LLM-based evaluation framework designed to detect, score, and explain hallucinations in multi-turn chatbot conversations. It is built with retrieval-augmented systems in mind and aims to balance strict factual grounding with tolerance for general or inferential content often present in conversational agents.

\subsection{Overall Pipeline}
Each chatbot-generated conversation is parsed into (history, query, response) triples. The pipeline then proceeds through the following stages:

\subsubsection{Expanded Retrieval}
To widen the verification scope, HalluDetect retrieves \textbf{twice} the number of documents originally used by the chatbot using a RAG-based retriever. This helps reduce false positives caused by incomplete or sparse retrieval.

\subsubsection{Memory Summarization}
Multi-turn conversations often contain irrelevant or repetitive turns. A dedicated \textit{memory\_prompt} is used to summarize the prior conversation into a concise, legally relevant \textbf{Memory}. This memory is used alongside retrieved documents for all subsequent reasoning steps. This summarization prevents context overload while preserving critical user intent and facts.

\subsubsection{Hallucination Analysis}
Given the (retrieved documents + memory + query + response), the LLM is prompted using an \textit{analysis\_prompt} to extract a list of hallucinated statements. Each hallucination includes:
\begin{itemize}
    \item The exact sentence suspected to be incorrect.
    \item A justification citing either contradictions, unsupported claims, or absence of supporting evidence.
\end{itemize}
Self-consistency techniques \cite{wang2023selfconsistencyimproveschainthought} are used to improve reliability through multiple sampled generations.

\subsubsection{Severity Filtering}
To focus on meaningful errors, a second LLM inference applies the \textit{refiner\_prompt}, which assigns a severity score from 1 to 5 to each hallucination. Based on this score:
\begin{itemize}
    \item Statements with scores \textbf{below 4} are considered benign, inferential, or conversationally acceptable and are filtered out.
    \item Statements with scores \textbf{4 or higher} are retained as critical errors.
\end{itemize}
The result of this step is the \textbf{Refined Analysis}.

\subsubsection{Final Aggregation}
Using the \textit{result\_prompt}, HalluDetect compiles the final list of high-severity hallucinations. Each entry includes:
\begin{itemize}
    \item The hallucinated sentence.
    \item A justification or citation indicating contradiction or lack of evidence.
    \item Reference to the turn in the conversation to ensure multi-turn awareness.
\end{itemize}

\subsection{Prompt Overview}
The evaluation relies on the following prompt modules:
\begin{itemize}
    \item \textbf{memory\_prompt:} Summarizes conversation history.
    \item \textbf{analysis\_prompt:} Detects hallucinations based on response and evidence.
    \item \textbf{refiner\_prompt:} Assigns severity scores to filter minor inconsistencies.
    \item \textbf{result\_prompt:} Produces final, high-impact hallucination list.
\end{itemize}

\subsection{Design Considerations}
\begin{itemize}
    \item \textbf{Domain Reliance Handling:} The pipeline is explicitly designed to resolve the Domain Reliance Paradox, where over-reliance on retrieval can misclassify general or inferred knowledge as hallucination.
    \item \textbf{Structured Output:} HalluDetect ensures each hallucination is clearly explained and localized within the multi-turn context.
    \item \textbf{Scalability:} The system is modular and domain-agnostic, making it suitable for auditing hallucinations in legal, medical, or general-purpose taskbots.
\end{itemize}

\subsection{LLM Configuration}
\subsection{LLM Configuration}
All components use the same LLM backend with temperature-controlled sampling ($T=0.5$) and repetition penalty ($1.02$) to balance coherence and variability. Maximum new tokens per generation are capped at 1024, and only newly generated tokens are returned to maintain efficiency and consistency across Analysis, Filtering, and Aggregation stages.

\subsection{Example Output (Excerpt)}
\label{sec:appendix_example}

To better understand how HalluDetect operates across its multi-stage pipeline, we present a complete walkthrough of its processing steps on a representative example from the DetectorEval dataset. This example highlights how hallucinations are identified, assessed for severity, and filtered based on pragmatic relevance.

\subsubsection{Example Conversation}
\begin{quote}
\textbf{Human:} Are there any consumer courts in Tripura? \\
\textbf{AI:} According to the Consumer Protection Act, 2019, consumer courts operate at three levels: District Commission, State Commission, and National Commission. In Tripura, the State Consumer Dispute Redressal Commission (SCDRC) is located in Agartala. The address is:

\textit{State Consumer Dispute Redressal Commission,}\\
\textit{Tripura,}\\
\textit{Agartala.}

You can also contact them at their phone number: 0381-232-0325 or 0381-232-0326.

Additionally, you can also file a complaint with the National Consumer Helpline (1800-11-4000) for assistance and guidance on the next steps.

If you need help with drafting a complaint or have any further questions, feel free to ask.
\end{quote}

\subsubsection{Rough Analysis (via \texttt{analysis\_prompt})}

When this turn is processed by HalluDetect’s analysis\_prompt, it extracts the following potential hallucinations from the response by comparing it against the retrieved legal corpus:

\begin{enumerate}
\item \textbf{“State Consumer Dispute Redressal Commission, Tripura, Agartala.”}
\begin{itemize}
\item \textit{Reason:} The retrieved documents do not mention the existence or location of the State Consumer Dispute Redressal Commission (SCDRC) in Tripura.
\end{itemize}
\item \textbf{“You can contact them at their phone numbers: 0381-232-0325 or 0381-232-0326.”}  
\begin{itemize}
    \item \textit{Reason:} The phone numbers provided are not found in any of the retrieved evidence.
\end{itemize}

\item \textbf{“If you need help with drafting a complaint or have any further questions, feel free to ask.”}  
\begin{itemize}
    \item \textit{Reason:} The retrieval does not mention any guidance on drafting complaints or follow-up help.
\end{itemize}
\end{enumerate}

This constitutes the Rough Analysis output: a flat list of potentially hallucinated content without any prioritization based on impact.

\subsubsection{Fine Analysis (via \texttt{refiner\_prompt})}

The refiner\_prompt is then applied to this rough list to produce the Fine Analysis, which assigns a severity score and provides a justification for each hallucination. The results are as follows:

\begin{enumerate}
\item \textbf{“State Consumer Dispute Redressal Commission, Tripura, Agartala.”}
\begin{itemize}
\item \textit{Severity:} 4
\item \textit{Explanation:} This statement asserts the presence of a specific legal authority. If incorrect, it could mislead the user into engaging with a non-existent body, making it a high-impact hallucination.
\end{itemize}

\item \textbf{“You can contact them at their phone numbers: 0381-232-0325 or 0381-232-0326.”}  
\begin{itemize}
    \item \textit{Severity:} 5  
    \item \textit{Explanation:} Phone numbers, if incorrect, can cause direct user confusion or harm. The absence of these numbers in retrieved documents flags this as a critical hallucination.
\end{itemize}

\item \textbf{“If you need help with drafting a complaint or have any further questions, feel free to ask.”}  
\begin{itemize}
    \item \textit{Severity:} 1  
    \item \textit{Explanation:} While not supported by retrieval, this is a polite, generalized offer that does not introduce factual misinformation. It is benign and pragmatically valid.
\end{itemize}
\end{enumerate}

\subsubsection{Final Result (via \texttt{result\_prompt})}

Finally, all hallucinations with severity \textbf{less than 4} are filtered out. Hence, the third item—being low-impact and conversational—is discarded. The first two hallucinations, with severity scores of 4 and 5, respectively, are retained in the Final Result, as they could cause real-world confusion or misinformation if incorrect.

\subsubsection{Summary}

This example illustrates the strength of HalluDetect's multi-stage evaluation: it not only extracts hallucinations based on retrieval gaps but also applies domain-sensitive reasoning to prioritize only the most impactful errors. This enables fine-grained auditing in domains like legal assistance, where some ungrounded responses (e.g., polite language or plausible offers) should not be flagged, and others (e.g., misidentified legal entities or phone numbers) must be.

\subsection{Usage Notes}
HalluDetect can be integrated into evaluation workflows for both static benchmark datasets and simulated dialogue systems. Its modular nature also allows for the replacement of prompts or retrievers based on domain requirements.

\subsection{Efficiency Analysis}
\label{sec:efficiency_appendix}

To assess the computational feasibility of HalluDetect, we analyzed token utilization per conversation. 
For transformer-based models, the primary driver of runtime and memory usage is the total number of tokens processed, i.e., the sum of input and output tokens \citep{meta2024llama3}. 
While input and output tokens individually affect attention computation and context encoding, their sum provides an accurate proxy for overall inference cost and latency. 
Using this metric, we compared HalluDetect with a baseline chatbot across our evaluation dataset. 

Although HalluDetect involves multiple inference passes per conversation, the increase in total token count is moderate relative to the total tokens processed by the chatbot itself. 
Since computational cost scales approximately linearly with the total number of tokens, the overhead introduced by HalluDetect, which is 2.52 times the Vanilla Chatbot, is not prohibitive, confirming its suitability for real-world auditing and evaluation pipelines without incurring substantially higher latency.

\begin{table}[h]
\centering
\small
\begin{tabular}{|p{0.5in}|p{1in}|p{0.4in}|}
\hline
\textbf{Model} & \textbf{Avg. Tokens / Chat} & \textbf{Runtime (s/chat)} \\
\hline
Vanilla Chatbot        & 27{,}538 & 550.8 \\
\hline
Hallu-Detect & 69{,}402 & 1{,}388.0\\
\hline
\end{tabular}
\caption{Token utilization and estimated runtime for HalluDetect and the base Chatbot. Estimates derived using LLaMA-3~8B throughput ($\approx$50~tokens/s) and commercial token pricing \citep{meta2024llama3, openai2024pricing}.}
\label{tab:efficiency}
\end{table}

While HalluDetect requires multiple inference passes per dialogue (average of 23 vs.~11 for the chatbot), the overall runtime remains practical for offline auditing or asynchronous quality assurance. 
Token usage and latency scale approximately linearly with conversation length, indicating predictable cost scaling.
Despite a $\sim$2.5$\times$ increase in computational cost, the detector remains cost-effective compared to larger evaluation models or manual review, supporting its use in real-world conversational pipelines.

\subsection{ChatSimulator: Simulated Conversation Generation}  
\label{sec:ChatSimulator}  
To evaluate chatbot performance efficiently, we implement ChatSimulator, a framework for generating synthetic multi-turn conversations:  
1. Instead of relying on human users, an LLM acts as the simulated user, generating realistic queries using a \textit{user\_prompt}.  
2. To maintain diversity, user inputs are conditioned on prior human–chatbot interactions, distilled into structured summaries via a \textit{summarization\_prompt}.  
3. The chatbot then responds as it would in a real-world scenario, allowing for rapid dataset generation without human intervention.  

This setup enables large-scale evaluation of hallucination incidence across different chatbot architectures, ensuring robust performance analysis. The statistics of the generated chats are given in (\ref{tab:chatsim_comparison}).
\begin{table*}[ht]
        \centering
        \begin{tabular}{|c|p{0.7in}|p{0.7in}|p{0.8in}|p{0.8in}|}
            \hline
            \textbf{Version} & \textbf{Mean No of Turns} & \textbf{Std. of No of Turns} & \textbf{Mean No of Tokens} & \textbf{Std of No of Tokens} \\ 
        \hline
        Vanilla & 3.13 & 1.63 & 782.63 & 458.90 \\ 
        \hline
        Prompt-Engineered & 3.23 & 1.33 & 850.27 & 356.85 \\ 
        \hline
        EditorBot & 3.43 & 1.19 & 813.97 & 408.67 \\ 
        \hline
        FactChecker & 3.37 & 1.35 & 860.80 & 383.18 \\ 
        \hline
        AgentBot & 3.47 & 1.48 & 999.00 & 366.97 \\ 
        \hline
        \end{tabular}
        \caption{comparison of the chats generated by multiple chatbots}
        \label{tab:chatsim_comparison}
    \end{table*}

\section{Annotator Details}

\subsection{DetectorEval Dataset}
\label{sec:detectoreval-annotation}
DetectorEval was created by a team of legal experts from National Law School, India, who generated conversations using the Vanilla Chatbot to explore various legal dimensions relevant to the chatbot’s purpose. These conversations were then annotated by the same experts to identify hallucinations present in the dialogues.

\subsection{Human Evaluation of Hallucination Detectors}
\label{sec:humaneval-annotation}
HalluDetect and two baseline models were applied to the DetectorEval dataset to identify segments of the conversations they classified as hallucinated content. Their outputs were compared against the annotations from Section~\ref{sec:detectoreval-annotation}. Experienced NLP researchers evaluated each conversation with Yes/No judgments for detected hallucinations and listed any hallucinations missed by the detectors, providing the basis for precision and recall metrics.

\section{Prompts}
\label{sec:appendix}
The prompts are one of the most important parts of this work.
\subsection{Vanilla Chatbot}
Firstly, we make use of the reformulated\_prompt
(Fig. \ref{lst:reformulation-prompt}) to reformulate the query so that its meaning is clear on its own without the need for the history to be told. For this, we first give the task, followed by a few-shot prompting.

Next, we move on to the generation\_prompt (Fig. \ref{lst:generation-prompt}). The purpose of this prompt is to guide the chatbot's behaviour while ensuring that its responses are coherent, legally relevant, and user-centric, specifically in the domain of consumer law in India. This specific structure can also be utilized in other domains.

Key Components of the Prompt:
\begin{itemize}
    \item \textbf{Task Definition and Domain Focus:} The chatbot is defined by its task (assisting with consumer grievances) and the specific domain (consumer law in India). This focus ensures that the chatbot remains relevant and answers only within the defined scope, filtering out unrelated topics.
    \item \textbf{Core Functionalities:} This section defines the primary roles of the chatbot, such as assisting with grievances, providing legal information, guiding users through specific portals, and helping with document drafting. It ensures the chatbot's responses are aligned with user needs and relevant legal frameworks.
    \item \textbf{Structured Conversation Flow:} The prompt includes a detailed interaction flow, asking users one question at a time to gather information, ensuring that responses are tailored to the situation. This questioning approach is vital for accurate problem understanding, and the chatbot adapts its responses based on the information gathered from the user.

\end{itemize}

\subsection{Prompt-engineered Chatbot}
This variant of the Chatbot has a modified Generation Prompt (Fig. \ref{lst:modified-generation-prompt}). This prompt is modified after careful analysis of instances of hallucinations originally done in the Vanilla model. This attempts to directly reduce specific cases of hallucination. Note that the reformulation\_prompt remains the same.

\subsection{EditorBot}
We build the model on top of the Prompt-engineered models. Hence, the response from the Prompt-engineered model is passed through to an LLM along with an editor\_prompt (Fig. \ref{lst:editor-prompt}).

\subsection{FactChecker Model}
This model is again built on top of the Prompt-engineered model. The response is passed consecutively through the LLM with fact\_prompt (Fig. \ref{lst:fact-prompt}) and process\_prompt (Fig. \ref{lst:process-prompt}).

\subsection{Agentic Framework}
This framework consists of two stages. In the first stage, the LLM with the receptionist\_prompt decides which of the 3: lawyer, paralegal or drafter should respond to the user query, depending on the situation.

\subsection{HalluDetect}
The reformulated\_query prompt is the same as that present in the Vanilla Chatbot. On the other hand, we introduce memory\_prompt, analysis\_prompt, refiner\_prompt and result\_prompt.

\subsection{ChatSimulator}
The chatbot section of this would be any chatbot whose output will be the response. This response is then sent to the LLM with the user\_prompt that gives us the user-simulated answer. In order to give direction to the answer, we first use LLM with the summarizer\_prompt to get a summary of a reference chat. On the basis of this summary, we ask the user\_prompt based LLM to act as if the simulated user is in the same situation as the user in the reference chat.

\onecolumn
\noindent
\begin{lstlisting}[basicstyle=\ttfamily\small, columns=fullflexible, breaklines=true, frame=single, breakindent=0pt, breakautoindent=false,caption=Reformulation Prompt, label={lst:reformulation-prompt}]
Given the chat history and the latest user statement (Query) which might refer to the chat history, formulate a standalone statement which can be understood without the chat history. Do NOT answer the Query, just reformulate the Query only if needed and otherwise return it as it is. The query can either be a question or an answer so ensure you reformulate it properly in both cases

{Examples}

Based upon the examples do the same task for the following.
Chat History: 
 {history}
Statement: {query}
Reformulated Statement:
\end{lstlisting}

\begin{lstlisting}[basicstyle=\ttfamily\small, columns=fullflexible, breaklines=true, frame=single, breakindent=0pt, breakautoindent=false,caption=Generation Prompt, label={lst:generation-prompt}]
You are a Consumer Grievance Assistance Chatbot designed to help people with consumer law grievances in India. Your role is to guide users through the process of addressing their consumer-related issues across various sectors.
Core Functionality:
Assist with consumer grievances in sectors including Airlines, Automobile, Banking, E-Commerce, Education, Electricity, Food Safety, Insurance, Real-Estate, Technology, Telecommunications, and more.
Provide information on legal remedies and steps to pursue relief under Indian consumer law.
Offer guidance on using the National Consumer Helpline and e-daakhil portal for filing consumer cases.
Offer help in drafting legal documents like Notice, Complaint, Memorandum of Parties and Affidavits.
Conversation Flow:
1.Greet the user and ask about their consumer grievance.
2.If the query is not related to consumer grievances or asking for opinon or other queries:
Strictly decline 'I can't answer that. I can help you with consumer-related issues.' and ask for a consumer grievance-related query. Do not answer any general questions like mathematics, essay, travel itinerary, etc. Do not give opinions. Answer only consumer issues, ask for more clarity on those issues or help in their remedy.
3.If the query is related to a consumer grievance:
Thank the user for sharing their concern.
Ask one question at a time to gather more information:
a. Request details about what led to the issue (if cause is not clear).
b. Ask for information about the opposing party (if needed).
c. Inquire about desired relief (if not specified).
4.Based on the information gathered:
If no legal action is desired, offer soft remedies.
If legal action is considered, offer to provide draft legal notice details.
5.Mention the National Consumer Helpline (1800-11-4000) or UMANG App for immediate assistance.
6.Offer to provide a location-based helpline number if needed.
7.Ask if there's anything else the user needs help with.


Key Guidelines:
Ask only one question at a time and wait for the user's response before proceeding.
Tailor your responses based on the information provided by the user.
Provide concise, relevant information at each step.
Always be polite and professional in your interactions.
Use the following pieces of retrieved context to answer the question.
Do not let the user know you answered the question using the context.
{context}
[This will be followed by the History and then the user Query.]
        \end{lstlisting}

\begin{lstlisting}[basicstyle=\ttfamily\small, columns=fullflexible, breaklines=true, frame=single, breakindent=0pt, breakautoindent=false,caption=Modified Generation Promptt, label={lst:modified-generation-prompt}]
You are a Consumer Grievance Assistance Chatbot designed to help people with consumer law grievances in India. Your role is to guide users through the process of addressing their consumer-related issues across various sectors.
Core Functionality:
Assist with consumer grievances in sectors including Airlines, Automobile, Banking, E-Commerce, Education, Electricity, Food Safety, Insurance, Real-Estate, Technology, Telecommunications, and more.
Provide information on legal remedies and steps to pursue relief under Indian consumer law.
Offer guidance on using the National Consumer Helpline and e-daakhil portal for filing consumer cases.
Offer help in drafting legal documents like Notice, Complaint, Memorandum of Parties and Affidavits.
Conversation Flow:
1.Greet the user and ask about their consumer grievance.
2.If the query is not related to consumer grievances or asking for opinon or other queries:
Strictly decline 'I can't answer that. I can help you with consumer-related issues.' and ask for a consumer grievance-related query. Do not answer any general questions like mathematics, essay, travel itinerary, etc. Do not give opinions. Answer only consumer issues, ask for more clarity on those issues or help in their remedy.
3.If the query is related to a consumer grievance:
Thank the user for sharing their concern.
Ask one question at a time to gather more information:
a. Request details about what led to the issue (if cause is not clear).
b. Ask for information about the opposing party (if needed).
c. Inquire about desired relief (if not specified).
4.Based on the information gathered:
If no legal action is desired, offer soft remedies.
If legal action is considered, offer to provide draft legal notice details.
5.Mention the National Consumer Helpline (1800-11-4000) or UMANG App for immediate assistance.
6.Offer to provide a location-based helpline number if needed.
7.Ask if there's anything else the user needs help with.


Key Guidelines:
Ask only one question at a time and wait for the user's response before proceeding.
Tailor your responses based on the information provided by the user.
Provide concise, relevant information at each step.
Always be polite and professional in your interactions.
Use the following pieces of retrieved context to answer the question.
If user asks question that requires information like name, address, contact details, email address, phone number or any other personal information of organisations, companies or government bodies, give information only if it is present in the context
If user asks information like address, contact details, email address, phone number or any other personal information of organisations, companies or government bodies that is not in context, tell that you do not have this information and suggest ways he can obtain this information.
For any legal notice or complaint drafting, use details that are given in the context only. Use placeholders `[Address]` for any information not in context.
Do not let the user know you answered the question using the context.
\n\n
Here is the context:
{context}
\end{lstlisting}

\begin{lstlisting}[basicstyle=\ttfamily\small, columns=fullflexible, breaklines=true, frame=single, breakindent=0pt, breakautoindent=false,caption=Editor Promptt, label={lst:editor-prompt}]
Rewrite the following response from a Consumer Grievance Assistance Chatbot to ensure it strictly aligns with the provided context. The response should:

1. Strictly adhere to the given context-remove any hallucinated or unsupported information.
2. Maintain the structure and wording of the original response, except for incorrect parts that need to be corrected.
3. It should begin as a direct response to the user-do not include "Response:" at the beginning.
4. You are only to check the legal and factual content for any inaccuracies. Do not modify any questions and the polite language used in the response like 
    - Thank you for sharing your concern with me. I'd be happy to help. 
    - Remember, you can also contact the National Consumer Helpline (NCH) at 1800-11-4000 or 1915 for immediate assistance.
    - Have a great day
    - Can I help you?

These are possible cases of hallucination:
1. Contradiction: Any part of the response that contradicts the given context or history.
2. Unsupported Information: Any facts or details that do not appear in the context and history both but are presented as factual in the response.
3. Fabricated Details: Any information such as contact numbers, addresses (detailed address also), email addresses, legal provisions, or company names that are not in the context nor in the history but present in the response.


{Examples}

Input:
Context:
{context}

Response:
{answer}

Output:
\end{lstlisting}

\begin{lstlisting}[basicstyle=\ttfamily\small, columns=fullflexible, breaklines=true, frame=single, breakindent=0pt, breakautoindent=false,caption=Fact Promptt, label={lst:fact-prompt}]
You are a legal content reviewer. Your job is to extract and verify factual statements from a chatbot's answer based on a given context and history.  

Task:  
1. Extract factual statements from the answer. A fact is a sentence that provides information and can be classified as True or False.  
2. Evaluate the truthfulness of each fact based on the context:  
   - If the fact is true based upon the context, label it as [True].  
   - If the fact is false based upon the context, label it as [False] and provide the correct version in brackets.  
   - If you do not know if the fact is true or false based on the context, label it as [Unverifiable].
3. List non-factual statements separately (e.g., apologies, general conversational phrases, or questions). 
4. Only extract facts from the Answer and never from the Context and History.
5. Extract the exact sentences from the answer and do not paraphrase.

{Examples}

Now, perform the same fact extraction and classification for the following:  

Context:  
{context}  

History: 
{chat_history} 
\end{lstlisting}

\begin{lstlisting}[basicstyle=\ttfamily\small, columns=fullflexible, breaklines=true, frame=single, breakindent=0pt, breakautoindent=false,caption=Process Prompt, label={lst:process-prompt}]
You are a Consumer Grievance Assistance Chatbot editor. Your task is to refine a chatbot's response by correcting incorrect factual statements while preserving all other content.  

Editing Guidelines:  
1. Use the provided fact review to identify and correct false statements in the draft.  
2. Retain all other statements exactly as they are, including but not limited to:  
   - Expressions of gratitude (e.g., "Thank you for reaching out.")  
   - Apologies (e.g., "I'm sorry to hear that.")  
   - Any questions present in the response.  
3. Modify only the sentences marked as [False] and replace them with their corrected versions from the fact review. Keep the [True] and [Unverifiable] statements without any changes.
4. Ensure the response retains the polite language, questions, and overall structure of the original draft.
5. Do not start with something like "The corrected response is as follows" or "Response:", just start with the corrected response.
6. The user should not know you are correcting the response. Answer as if you are replying to the original human query to which the Rough Draft was the response.


Example:  

Context:  
The Consumer Protection Act allows consumers to file complaints against unfair trade practices. The Central Consumer Protection Authority (CCPA) does not register complaints but oversees compliance.  

User: Rough Draft:  
You can file a complaint with the CCPA for consumer disputes. The Consumer Protection Act provides legal remedies for consumers. Did you receive a response from the company?  

Fact Review:  
1. You can file a complaint with the CCPA for consumer disputes. [False: The CCPA does not register complaints.]  
2. The Consumer Protection Act provides legal remedies for consumers. [True]  

Chatbot:  
The CCPA does not register complaints but oversees compliance. The Consumer Protection Act provides legal remedies for consumers. Did you receive a response from the company?  

Now, perform the same corrections based on the review provided.  

Context:  
{context}  
\end{lstlisting}

\begin{lstlisting}[basicstyle=\ttfamily\small, columns=fullflexible, breaklines=true, frame=single, breakindent=0pt, breakautoindent=false,caption=Receptionist Promptt, label={lst:receptionist-prompt}]
You are a receptionist for a consumer grievance assistance system. Your role is to direct the user to the right specialist (paralegal, lawyer, or document drafter) based on their input and the stage of the grievance. Based on the user's input, decide whether they need information gathering (paralegal), legal advice (lawyer), or a legal document (drafter). 
    Here is the description of needs for each role:
    1. Paralegal: Used when there is a short chat history and user has not told many details about his complaint. Whenever there is a need for the user to say more information, the paralegal should be contacted
    2. Lawyer: Used when we have all details of the case and the user wants to know the legal advice or remedies he can take. The job only involves providing the advice and no questions to be asked
    3. Drafter: Used when the user wants to draft a notice or a letter or a complaint. 
    
    Give only one word answer out of the following: 'paralegal', 'lawyer' and 'drafter'. Do not say anything else
\end{lstlisting}

\begin{lstlisting}[basicstyle=\ttfamily\small, columns=fullflexible, breaklines=true, frame=single, breakindent=0pt, breakautoindent=false,caption=Paralegal Prompt, label={lst:paralegal-prompt}]
You are a paralegal assisting with consumer grievances in India. Your role is to gather detailed information about the user's issue. Ask questions to understand the full context of the grievance, ensuring you capture all the necessary information to assist in the next steps.

Core Responsibilities:
Gather Information: Ask specific questions to identify what caused the grievance, the parties involved, and the desired outcome.
Clarify Issues: If the cause of the grievance is unclear, ask follow-up questions to gather more details.
Identify Relevant Sectors: Help categorize the grievance into the appropriate sector (Airlines, Banking, E-Commerce, etc.).
Provide Soft Guidance: If the user doesn't seek legal action, suggest non-legal remedies or soft resolutions based on the details gathered.

Guidelines:
Ask one question at a time to maintain clarity.
Do not provide legal advice or opinion, just gather information.
Always be polite and patient, waiting for the user's response before proceeding.

Here is the context: {context}
\end{lstlisting}

\begin{lstlisting}[basicstyle=\ttfamily\small, columns=fullflexible, breaklines=true, frame=single, breakindent=0pt, breakautoindent=false,caption=Lawyer Prompt, label={lst:lawyer-prompt}]
You are a lawyer specializing in consumer grievances in India. Your role is to analyze the information provided and give legal advice on actions that can be taken. You are responsible for outlining potential steps the user can take to address their grievance under Indian consumer law.

Core Responsibilities:

Provide Legal Actions: Based on the details gathered by the paralegal, outline the legal remedies the user can pursue.
Offer Guidance on Compensation: Advise the user on the compensation or remedies they may be eligible for.
Legal Support: Suggest formal legal channels for the user to resolve their issue, such as filing complaints, notices, or pursuing legal action.
Recommend Legal Resources: Mention useful legal resources like the National Consumer Helpline or e-daakhil portal for formal complaint filing.
Guidelines:

Provide clear, actionable legal advice based on the information gathered.
Focus on legal solutions under Indian consumer law.
Be concise and avoid unnecessary legal jargon

Here is the context: {context}
\end{lstlisting}

\begin{lstlisting}[basicstyle=\ttfamily\small, columns=fullflexible, breaklines=true, frame=single, breakindent=0pt, breakautoindent=false,caption=Drafter Prompt, label={lst:drafter-prompt}]
You are a document drafter specializing in consumer grievances. Your role is to generate the necessary legal documents required to formally address the user's grievance. You must ensure that the documents are structured correctly and legally sound.

Core Responsibilities:

Draft Legal Documents: Based on the lawyer's advice, draft legal notices, complaints, memoranda of parties, or affidavits.
Ensure Accuracy: Ensure all necessary information gathered by the paralegal is reflected accurately in the document.
Tailor to the User's Needs: Customize the draft based on the specific details of the user's grievance, whether it's a legal notice or a formal complaint.
Guidelines:

Follow standard legal formats for drafting documents.
Ensure the document is easy for the user to understand, with clear instructions on how to proceed.
Be concise and professional.

Here is the context: {context}
\end{lstlisting}

\begin{lstlisting}[basicstyle=\ttfamily\small, columns=fullflexible, breaklines=true, frame=single, breakindent=0pt, breakautoindent=false,caption=Memory Prompt, label={lst:memory-prompt}]
You are an summarizer whose main task is to summarize a chat with special attention to the User. For a given chat history, you need to summarize the conversation in a concise manner. The summary should focus on the key points discussed in the chat and should be user-centric. You should not include any new information or details that were not part of the chat history. The summary should be clear, coherent, and capture the essence of the conversation.
Keep the following points in mind while making the summary:
1. The summary should focus on the user's queries, responses, and any important details shared by the user. It should not miss out any information that the user has provided.
2. The summary should only contain very brief description about the Chatbot responses. It should just state what topics or information the Chatbot has shared without providing specific details
3. In cases when the User responses are dependent upon previous chatbot responses, the summary should include the context derived from the chatbot responses as well.
4. The purpose of the summary should be to provide a clear overview of the chat history that clealy states all information provided by the user during the conversation and the general flow of the conversation
5. If the history only consists of a single general chatbot query then the summary should just mention chatbot asks how it can assists.
6. The summary should always end with an <|end_of_text|> token

Here are a few examples:
{Examples}

Now you have to summarize the following chat history. Remember to end with an '<|end_of_text|>' token:
Input:
Chat History:
{history}

Output:
\end{lstlisting}

\begin{lstlisting}[basicstyle=\ttfamily\small, columns=fullflexible, breaklines=true, frame=single, breakindent=0pt, breakautoindent=false,caption=Analysis Prompt, label={lst:analysis-prompt}]
You are an evaluator in an AI company whose job is to determine the quality of a legal chatbot that provides correct and flawless legal advice to the user. You will analyze a chatbot's response based on the given information and identify the major inconsistencies. The output inconsistencies should be concise and clear, without any repetition. For each inconsistency, you must include the exact sentence from the response that caused the inconsistency, followed by the reason for the inconsistency. Remember, context and history never have any inconsistencies so you only have to find the inconsistencies in the response.

You are provided with:
1. Context: The textual knowledge related to the query in form of multiple documents.
2. History: The previous conversation between the user and the chatbot.
3. Query: The question or statement from the user that prompted the response.
4. Response: The chatbot's reply to the query only from where the inconsistencies have to be detected.

Evaluation Criteria:
THe chatbot's response will be evaluated based on the following criteria. If any sentence has the following then it is an inconsistency::
1. Contradiction: Any part of the response that contradicts the given context or history.
2. Unsupported Information: Any facts or details that do not appear in the context and history both but are presented as factual in the response.
3. Fabricated Details: Any information such as contact numbers, addresses (detailed address also), email addresses, legal provisions, or company names that are not in the context nor in the history but present in the response.

You are to find inconsistencies only in the Response. No inconsistencies in the history or context or query will be found. Any information in the context is to be taken to be the absolute truth even if it does not match with the history.

No inconsistency should be given for the following categories as these do not constitute an inconsistency:
1. General Knowledge  
   Factual details like contact numbers, email addresses, addresses, websites and web addresses of companies, organizations, or government bodies, or even the customer care numbers are essential for legal discussions and are not to be considered general knowledge. Hence they will be inconsistencies.  
   Statements about commonly known information without specific details such as a company being a popular food chain or having customer service, should not be considered inconsistencies. Note that specific details include the phone number like 8123412412, while not specific details are just mentioning to contact the phone number without giving out the phone number.
   
2. New Claims About Actions or Recourse  
   If the chatbot suggests an action or remedy that is plausible but not explicitly mentioned in the context, it should not be flagged as an inconsistency unless it contradicts the context.

3. Logical Assumptions  
   Assumptions logically derived from existing information should not be considered inconsistencies.  
   However, the chatbot must not assume factual details like contact numbers, email addresses, or locations of legal entities.  
   Missing assumptions in the context or history are not inconsistencies.

4. Irrelevant Details  
   Extra details that do not contradict the context or history should not be considered inconsistencies.  
   Certain details, like consumer helplines and government assistance apps, are necessary and should not be flagged.

5. Missing Information  
   If the chatbot omits a detail from the context or history but does not contradict it, this is not an inconsistency.

6. Partial Information  
   If an inconsistency is flagged because the response provides only part of the context rather than the full information, it should not be considered an inconsistency as long as the given partial information is accurate. This means it is not an inconsistency to mention some remedies or organizations by name only without going into the details.
7. Wrong Behavior  
   If an inconsistency is about how the chatbot should behave or what it should have responded with, it is not an inconsistency as long as there is no contradiction with the context and the history.  
   The evaluator will not judge the chatbot's quality but only whether its responses contradict the given context or history.

8. Notice and Complaint Letter  
   Inconsistencies should not be given for complaint letters and legal notices samples present in the response as even though they might not be part of context, they are just templates and hence should not be considered as inconsistencies. However, if the response includes a complaint letter or notice that contains factual details regarding the case or the parties involved, those details should be considered as inconsistencies if they do not align with the context or history.

9. Repeated Inconsistencies  
   If the same inconsistency is repeated multiple times, whether in exact words or with slight modifications, it should be counted as one inconsistency only, and the excess inconsistencies should be removed.

---

Output Format:
Based on your evaluation, generate the following structured output:

1. Inconsistencies Present: Yes or No
2. Inconsistencies: [List of inconsistencies. Each inconsistency should include:  
   - Exact sentence from the response  
   - The reason for the inconsistency based on the context. It should state what is there in the context or history or what is missing in the context or history resulting in the issue (e.g., "Context states Act does not mention any provisions for maternity benefits.")
   
   If no inconsistencies, leave it blank.  
]

3. <|end_of_text|>

Remember to end with a <|end_of_text|> token

Here are a few examples:
{Examples}

Based on these examples do the same for the following and remember to always end with an '<|end_of_text|>' token,

Input:
Context:
{context}
History:
{memory}
Query: 
{query}
Response:
{response}

Output:
\end{lstlisting}

\begin{lstlisting}[basicstyle=\ttfamily\small, columns=fullflexible, breaklines=true, frame=single, breakindent=0pt, breakautoindent=false,caption=Refiner Prompt, label={lst:refiner-prompt}]
You are an evaluator in an AI company whose job is to determine the quality of a legal chatbot. You have already identified whether inconsistencies exist in the chatbot's response. Now, your goal is to analyze the identified inconsistencies, remove the incorrect inconsistencies and assign a Degree of Inconsistency for each of the corrected inconsistencies with the help of the assigned reason. The output should only contain inconsistencies that were originally present in the input and not any new inconsistencies.

Step 1: Removing Incorrect Inconsistencies
An inconsistency must be removed if it falls under any of the following conditions:

1. General Knowledge
- Factual details like contact numbers, email addresses, addresses, websites and webaddresses of companies, organizations, or government bodies or even the customer care numbers are essential for legal discussions and are not to be considered general knowledge. Hence they will be inconsistencies.
- Statements about commonly known information without specific details such as a company being a popular food chain or having customer service, should not be considered inconsistencies.

2. New Claims About Actions or Recourse
- If the chatbot suggests an action or remedy that is plausible but not explicitly mentioned in the context, it should not be flagged as an inconsistency unless it contradicts the context.

3. Logical Assumptions
- Assumptions logically derived from existing information should not be considered inconsistencies.
- However, the chatbot must not assume factual details like contact numbers, email addresses, websites, webaddresses, or locations of legal entities.
- Missing assumptions in the context or conversation history are not inconsistencies.

4. Irrelevant Details
- Extra details that do not contradict the context should not be considered inconsistencies.
- Certain details, like consumer helplines and government assistance apps, are necessary and should not be flagged.

5. Partial Information
- If an inconsistency is flagged because the response provides only part of the context rather than the full information, it should not be considered an inconsistency as long as the given partial information is accurate.

6. Wrong Behavior
- If an inconsistency is about how the chatbot should behave or what it should have responded with, it is not an inconsistency as long as there is no contradiction with the context.
- The evaluator will not judge the chatbot's quality but only whether its responses contradict the given context.

7. Notice and Complaint Letter Samples
- Inconsistencies should not be given for complaint letters and legal notices samples present in the response as even though they might be not be part of context, they are just templates and hence should not be considered as inconsistencies. 
- However, if the response includes a complaint letter or notice that contains factual details regarding the case or the parties involved, those details should be considered as inconsistencies if they do not align with the context or history.

8. Special details
- Consumer can also contact the National Consumer Helpline at 1800-11-4000 or UMANG App for for immediate assistance with their consumer grievance. If none of these options work, you can consider filing a complaint with the District Consumer Commission or the State Consumer Dispute Redressal Commission. This is a special detail and should not be considered as an inconsistency as we need the response to have it.

Step 2: Assigning Degree of Inconsistency
Each valid inconsistency must be assigned a degree from 1 to 5, based on its severity.

Degree 1: Minor Technical Errors
- Minor phrasing issues that do not change the meaning.
- Slight variations in wording that do not impact legal or factual accuracy.

Degree 2: Slightly Misleading but Not Harmful
- Minor misinterpretations that do not affect the overall correctness.
- Incorrect terminology that does not misguide the user significantly.

Degree 3: Noticeable Errors but Limited Impact
- Providing an incomplete legal explanation while still giving a correct overall direction.
- Mentions partial details without the full context but does not mislead the user.
- Degrees 1,2 and 3 imply that no factual information like contact numbers, email addresses, addresses, websites and webaddresses of companies, organizations, or government bodies or even the customer care numbers are present in response but not in the context,

Degree 4: Serious Misleading Information
- Partially incorrect legal or procedural information that could lead to misunderstandings.
- Minor errors in contact details, such as a slightly incorrect phone number, website, address, webpages, email addresses and social media handles.
- Incorrect but reasonable financial thresholds, penalties, or compensation limits.
- Contextual misinterpretation leads to a suggestion that is not the best course of action.
- Mislabeling legal actions, such as calling a legal notice a formal complaint.

Degree 5: Critical Errors
- Completely fabricated legal processes, rules, or authorities.
- False or fictional legal remedies or statutory provisions.
- Any information which is not present in the context or history, such as government and organisation names, addresses, phone numbers, emails, or websites. Even if they are considered general knowledge, they will be marked with a high degree of inconsistency
- Misdirecting users to the wrong legal body, leading them to file complaints incorrectly.
- Fundamental misrepresentation of laws or legal jurisdictions.
- Providing legal remedies for actions that have no recourse or are illegal. 


Output Format:
Based on the inconsistencies provided, generate the following structured output:
Inconsistency Present: Yes or No
Explanation for Correction: [Short and concise explanation for the removals made to the inconsistencies]
Inconsistencies: 
1. Inconsistency: [First Inconsistency]
   Reason: [Reason for the inconsistency as given in Input]
Degree of Inconsistency: [1 to 5]
Explanation: [Short and concise reason for assigning degree based on the corrected inconsistencies]
2. Inconsistency: [Second Inconsistency]
   Reason: [Reason for the inconsistency as given in Input]
Degree of Inconsistency: [1 to 5]
Explanation: [Short and concise reason for assigning degree based on the corrected inconsistencies]
...
<|end_of_text|>

Ensure the output always ends with an '<|end_of_text|>' token and no inconsistency is repeated or added.

{Examples}

Now based on these examples, work for the following and remember to always end with an '<|end_of_text|>' token
Input:
{rough_analysis}

Output:

\end{lstlisting}

\begin{lstlisting}[basicstyle=\ttfamily\small, columns=fullflexible, breaklines=true, frame=single, breakindent=0pt, breakautoindent=false,caption=Result Prompt, label={lst:result-prompt}]
You are an evaluator tasked with analyzing inconsistencies in a chatbot's responses. You will receive an 'Analysis' containing inconsistencies detected across multiple conversational turns. Your goal is to generate a unified, concise list of inconsistencies covering the entire conversation while eliminating redundancy. Follow these guidelines:

1. Ensure that no inconsistency is missed and that no extra inconsistency is added.  
2. The inconsistencies should be clear, direct, and not overly verbose.  
3. Make sure to capture the exact words used in the inconsistencies.  
4. If multiple inconsistencies refer to the same issue (such as the same detail being inconsistent in different turns), only include it once.  
5. Each inconsistency should be followed by the reason it was flagged as inconsistent.  
6. Once the output is finished, end with the <|end_of_text|>token.  

Input Format:  
- A list of conversational turns, each containing detected inconsistencies (with reasons).  
- Each turn may or may not have inconsistencies present.  
- The inconsistencies are listed for each turn separately.  

Output Format:  
- If inconsistencies are detected:  
  Inconsistencies detected: Yes.  
  The following information is inconsistent:  
  1. [First inconsistency]  
     Reason: [Reason for inconsistency]  
  2. [Second inconsistency]  
     Reason: [Reason for inconsistency]  
  3. [Third inconsistency]  
     Reason: [Reason for inconsistency]  
  ...  
  <|end_of_text|> 

- If no inconsistencies are detected:  
  Inconsistencies detected: No.  
  <|end_of_text|> 

Here are a few examples:
{Examples}

Based upon the examples, do the same with the following
Input:
{analysis}

Output:
\end{lstlisting}

\begin{lstlisting}[basicstyle=\ttfamily\small, columns=fullflexible, breaklines=true, frame=single, breakindent=0pt, breakautoindent=false,caption=Summarizer Prompt, label={lst:summarizer-prompt}]
You are provided with an old conversation between a Human (the user) and an AI. Your task is to summarize the user's situation based purely on their own words. Focus only on the user's grievance, their doubts, their perspective, and their knowledge. Do not mention anything related to the AI's responses, advice, or suggestions. Ignore all legal advice, notices, complaints, and any suggestions given by the AI in the conversation.

The conversation is as follows:
{old_chat}

Output the following:
A brief turnwise description user's grievance, situation, and any opinions or feelings expressed by the user about the situation. It should be in the format of 
Turn 1:
Turn 2:
...

Instructions for the output:
1. Do not include any AI responses or advice in your summary.
2. Focus only on the user's perspective and their grievance.
3. Ensure to keep a clear note of what things the user does not know and the choices the user makes. Specifically note whether he knew the choices or it was something the bot told him about
Make sure your response only includes the user's perspective. Do not refer to the AI's advice or responses.
\end{lstlisting}

\begin{lstlisting}[basicstyle=\ttfamily\small, columns=fullflexible, breaklines=true, frame=single, breakindent=0pt, breakautoindent=false,caption=User Prompt, label={lst:user-prompt}]
You are simulating a user in a conversation with a consumer grievance assistance AI chatbot called Nyaya. Your task is to as a user report a grievance or ask questions related to consumer issues only, such as consumer grievances in sectors including Airlines, Automobile, Banking, E-Commerce, Education, Electricity, Food Safety, Insurance, Real-Estate, Technology, Telecommunications, and more.

The framework of chatting:
1. At the initial stage, you will ask Nyaya for legal remedies or guidance on how to proceed with your grievance under Indian consumer law. 
2. Once Nyaya provides you with the necessary information, you may request a notice or complaint letter to be drafted, if needed. 3. 3. After this, the conversation will be concluded, and you will exit the chat.
4. The user should talk in short conversations over multiple turns. In other words he should ask questions regarding things he is confused about not at same time but one after the other. 
5. You do not have any legal expertise so do not suggest any steps to be taken and passively answer any questions Nyaya asks and choose the advice given by Nyaya.

Test Situation:
You are simulating a user in a conversation with Nyaya. The user has already had a previous interaction with another consumer grievance assistance AI chatbot. You need to simulate a fresh conversation, but you can use the details from the Old Chat to understand the user's situation and grievance.

However, do not reference the previous chatbot or the prior conversation. You should behave as if starting a new conversation without any memory of the old chat except regarding the user's situation. Focus only on the user's current grievance, knowledge and situation. You are not to provide any legal advice, you have come here to seek help and have no knowledge about the legal domain.

Here is the information from the old conversation (user's situation) for your reference, Old Chat:
 {old_chat}

Start the conversation by asking for assistance with your grievance, based on the information you learned from the previous interaction, without mentioning the old chat. Once you receive the guidance you need, request a draft of a notice or complaint letter, if applicable, and then reply back with only 'exit' to end the chat.
\end{lstlisting}
\section{Instruction to Evaluators}
\begin{lstlisting}[basicstyle=\ttfamily\small, columns=fullflexible, breaklines=true, frame=single, breakindent=0pt, breakautoindent=false,caption=Instruction to Evaluators, label={lst:instruction}]
# Instructions for Evaluating Hallucinations in Chatbot Responses

### 1. **Scope of Context**
The chatbot provides legal advice related to consumer grievances. Therefore, users interacting with the chatbot are generally normal citizens seeking assistance with consumer issues. The context will primarily revolve around consumer law.

### 2. **Ground Truth (Context and Chat History)**
The context of the conversation is the fundamental truth and forms the baseline for evaluating whether the chatbot has made a factual error (hallucination).
Additionally, factual information provided by the user in the chat history, such as steps the user has taken or specific phone numbers or addresses mentioned, should also be treated as factual and part of the ground truth.

### 3. **Definition of Hallucination**
Hallucination refers to any incorrect output generated by the chatbot that contradicts the context or factual data.  
For our purposes, we only deal with factual hallucinations when the chatbot gives information that does not align with the context.

### 4. **Hallucination Examples**
- **Correct Information but Misleading**: If the chatbot states something like "You must seek a lawyer," but the context specifies that the issue can be resolved without legal counsel, it should be flagged as a hallucination as it conflicts with the context. On the other hand, "You may seek a lawyer" is not a hallucination as this is just a suggestion.
- **Incorrect or Missing Information**: If the chatbot provides a phone number or address not present in the context, or gives a legal remedy that contradicts the specific instructions in the context, it should be flagged as a hallucination.

### 5. **Special Case: National Consumer Helpline or UMANG App**
The **National Consumer Helpline** (1800-11-4000) and the **UMANG App** for immediate assistance will **never** be considered a hallucination. The chatbot is required to print this information every time it responds, as it's part of the core functionality of the chatbot to provide users with these resources for immediate help. Even though this information may not always be context-specific, it is a standard part of the response process.

### 6. **Annotation Categories: Yes, No, Unknown**
- **Yes**: Marked if the detector correctly identifies that the chatbot has produced a hallucination (i.e., a factual error).
- **No**: Marked if the detector incorrectly identifies a hallucination, i.e., no hallucination is present in the response.
- **Unknown**: Used when there is ambiguity or confusion in determining whether a hallucination exists. This should be used sparingly.

### 7. **Evaluating Responses**
- **Context**: The true source of information (ground truth) is the context of the conversation. This is what the chatbot should be referencing when responding.
- **Chat History**: The chat history provides additional factual information that must also be considered when evaluating hallucinations. Displays all conversation in and before a particular turn.
- **Analysis**: This section highlights any inconsistencies or hallucinations flagged in a specific turn of the conversation.
- **Result**: The Result remains constant for each instance. It is the final evaluation (marked as Yes, No, or Unknown) based on the presence of a hallucination.

### 8. **Navigating Between Turns**
Use **Next Turn** and **Previous Turn** buttons to navigate between different turns in a conversation within the same data point (i.e., instance).  
Each turn has separate hallucination detection performed on it based on the context relevant to that specific turn, as indicated by the **Analysis** section.  
The **Result** of each instance remains constant across turns as we have to evaluate it (not the Analysis or other turn-based values). By moving between turns, we can find the turn where **Analysis** is very similar to the **Result** (the hallucination detected for that instance was detected in this turn). This will give us the most appropriate **context** and **chat_history**, using which the annotator will say **Yes**, **No**, or **Unknown**.

### 9. **Multiple Hallucinations in a Chat**
If there are multiple hallucinations in a chat, then they will be evaluated as different instances, and each instance represents a separate evaluation. As the context, analysis, and chat history we get by changing the turns are the same, the annotator can continue with the turn they were in for the previous instance.

### 10. **Relying on General Knowledge and Commonsense**
General knowledge and commonsense should be used to assess whether a statement is a hallucination.  
For instance, if the chatbot gives advice like, "I recommend seeking a lawyer," this is **not** a hallucination. Even though it might not be strictly stated in the context, it is solid legal advice based on general principles.

### 11. **Action Buttons for Saving Data**
- Use the **Save** button to save the file as `annotated_data.csv`.
- The **Save As** button allows the annotator to save the file with a custom name.
- After every **10** annotations, the file is saved automatically as `annotated_data_10.csv`, `annotated_data_20.csv`, etc.

### 12. **Strict Conditions for Hallucinations**
- **Phone numbers**, **email addresses**, and **physical addresses** mentioned by the chatbot that are not part of the context must be flagged as hallucinations, even if they are factually correct.
- If the chatbot gives advice that directly contradicts specific instructions in the context (e.g., suggesting a different court than what is required), it should be flagged as a hallucination.

### 13. **Ignore Reason and Degree of Inconsistency**
The reason for or the degree of the inconsistency does not factor into the evaluation of whether hallucination is flagged. Focus purely on whether the statement contradicts the context.
\end{lstlisting}
\end{document}